\documentclass{article}
\usepackage[utf8]{inputenc}
\usepackage{amsmath, amssymb}
\usepackage{graphicx}
\usepackage{hyperref}
\usepackage{authblk}
\usepackage{wrapfig}
\usepackage{float}
\usepackage{booktabs}

\usepackage{tikz}
\usetikzlibrary{arrows.meta, positioning, shapes.geometric, calc}

\title{Recurrent Memory-Augmented Transformers with Chunked Attention for Long-Context Language Modeling}

\author{Ankit Kashyap}
\affil{Independent Researcher, Patna, India\\
\texttt{ankitchaahat2001@gmail.com}}

\date{July 1, 2025}

\begin{document}

\maketitle

\section*{Abstract}
\noindent
Large Language Models (LLMs) have significantly advanced the state-of-the-art in natural language processing, achieving remarkable performance in tasks such as machine translation, code generation, document summarization, and dialogue modeling. Despite their success, a major limitation of existing LLMs lies in their fixed-length context windows, which restrict the model's ability to retain and reason over long-range dependencies. This bottleneck is particularly evident in domains requiring deep contextual understanding, such as multi-turn dialogue, long document analysis, and sequential code completion.

To address this, we present a novel Transformer-based architecture that augments standard self-attention with two additional mechanisms: \textbf{localized chunked attention} and a \textbf{learnable recurrent memory module}. These three attention paths are integrated into a unified hybrid attention block, enabling the model to capture both fine-grained local patterns and long-range cross-sequence dependencies. Unlike conventional designs, our memory module supports \textbf{gated memory updates} inspired by recurrent networks and follows a \textbf{FIFO-style memory bank management} strategy using efficient tensor operations. This design allows the model to retain a persistent, dynamically updated memory across input sequences, without growing the attention cost quadratically.

In addition, we implement \textbf{Rotary Positional Encoding (RoPE)} at the \textit{multi-head level}, ensuring consistent and scale-invariant positional embeddings across heads — an enhancement rarely implemented manually in prior work. Our architecture also incorporates \textbf{Pre-LayerNorm residual connections} to improve training stability in deep model stacks and is developed \textbf{entirely from scratch in PyTorch}, avoiding any dependency on pre-built transformer libraries such as Huggingface or Fairseq. 

\textbf{Unlike prior architectures such as Transformer-XL and Longformer, our design combines gated FIFO memory, multi-path hybrid attention (full, chunked, and memory), and per-head RoPE into a unified lightweight framework that is both flexible and efficient for long-context modeling.}

\section{Introduction}
\noindent
Large Language Models (LLMs) such as GPT, BERT, and their derivatives have demonstrated impressive capabilities across a variety of natural language understanding and generation tasks. These models are typically based on the Transformer architecture, which uses self-attention to learn dependencies across sequences. Despite their success, Transformers remain fundamentally constrained by a fixed-length context window, which limits their ability to model long-term dependencies in input sequences. This limitation is particularly problematic in real-world applications that involve long-form text, such as legal document processing, multi-turn dialogues, and code modeling, where important contextual information may lie far apart in the token stream.

Traditional solutions, such as increasing the input context length or using hierarchical segmentations, suffer from either exponential compute and memory costs or degradation in contextual coherence. To address this, several research efforts have explored augmenting Transformers with external memory mechanisms \cite{transformerxl}, sparse and local attention variants \cite{longformer, bigbird}, and retrieval-based architectures \cite{retro}. While these architectures have shown promise, they often come with increased architectural complexity, inference latency, or reliance on pretraining tricks and large-scale infrastructure.

In this work, we present a lightweight yet effective solution: a Transformer architecture augmented with a combination of \textbf{full-sequence self-attention}, \textbf{localized chunked attention}, and a \textbf{learnable recurrent memory module}. This unified attention framework enables the model to simultaneously capture short-range patterns within chunks and maintain long-term dependencies across sequences. Our design includes a gated memory update mechanism, inspired by recurrent networks, and a \textbf{FIFO-style memory bank} to simulate continuous contextual learning without unbounded memory growth. This approach allows the model to retain relevant past information and reuse it during future computations.

We also adopt \textbf{Rotary Positional Encodings (RoPE)} at the multi-head level, enhancing the model’s ability to generalize across positions with scale-invariant structure. Furthermore, we incorporate \textbf{Pre-LayerNorm residual connections}, a proven technique to improve training stability in deep transformers. All modules are implemented \textbf{entirely from scratch in PyTorch}, without relying on external transformer libraries such as Huggingface or Fairseq, making the architecture transparent, interpretable, and easily extensible for research purposes.

Our approach not only bridges the gap between full-attention and memory-augmented Transformers but also provides a practical solution for resource-constrained training and inference scenarios. We empirically demonstrate improved long-context retention and reduced memory overhead using synthetic and benchmark datasets. The model achieves competitive perplexity with significantly fewer parameters and simpler architecture compared to conventional long-context models.

\textbf{Our key contributions can be summarized as follows:}
\begin{itemize}
    \item We propose a hybrid attention mechanism that fuses full self-attention, chunked attention, and memory attention into a single transformer block.
    \item We introduce a novel gated recurrent memory module with a rolling FIFO structure to simulate long-range memory.
    \item We apply Rotary Positional Encoding at the per-head level to improve position generalization across attention heads.
    \item We provide a fully modular, from-scratch PyTorch implementation suitable for educational and applied research.
    \item We demonstrate promising empirical results on long-context tasks with reduced memory cost and strong generalization.
\end{itemize}

The rest of the paper is organized as follows: Section~\ref{sec:related} covers related work. Section~\ref{sec:method} describes the architecture in detail. Section~\ref{sec:conclusion} concludes with discussions and future directions.

\section{Model Architecture}
\label{sec:architecture}

\begin{wrapfigure}[14]{r}{0.42\textwidth}
    \centering
    \vspace{-10pt}
    \includegraphics[width=0.40\textwidth]{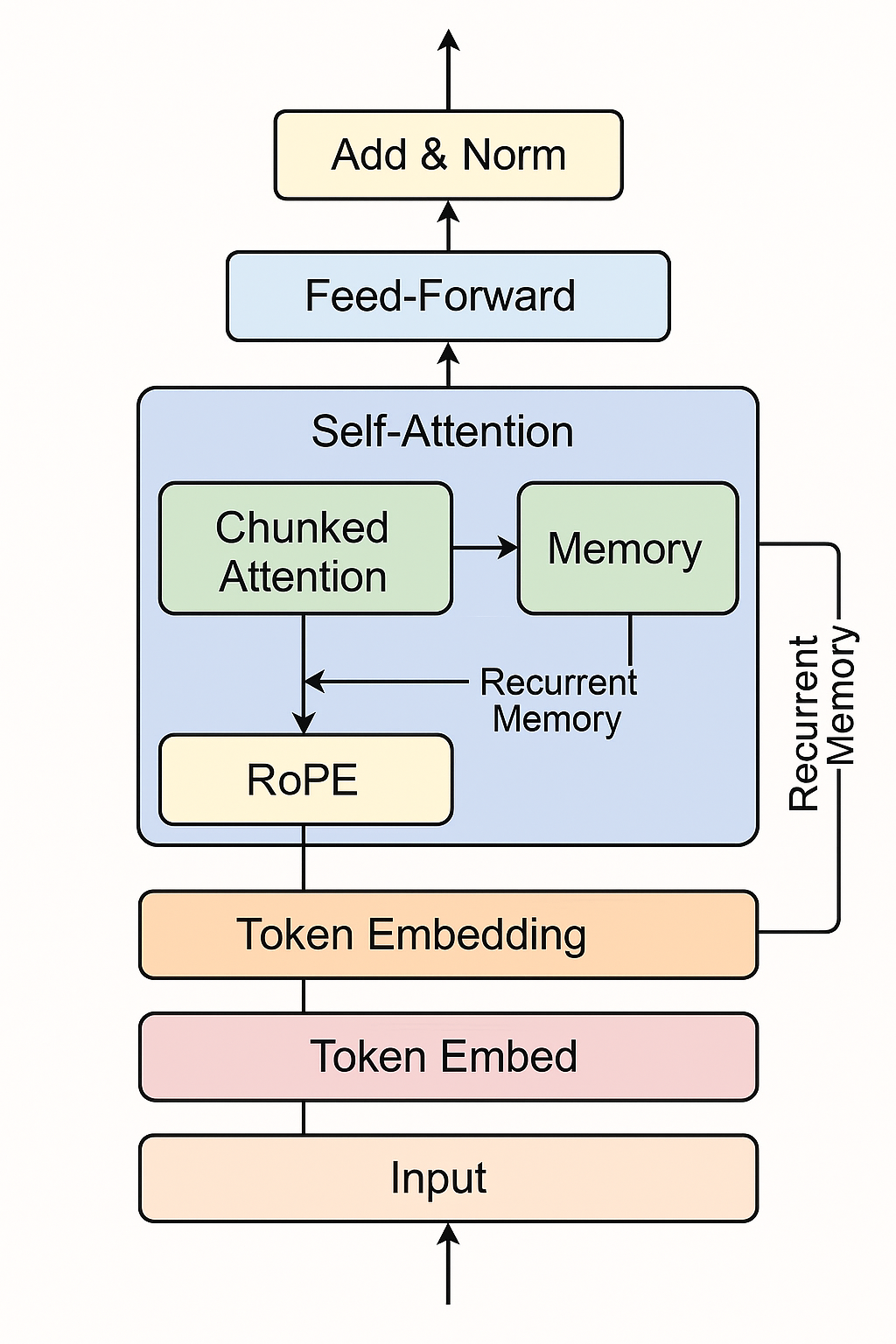}
    \caption{Overview of the proposed hybrid Transformer architecture combining chunked attention, recurrent memory, and RoPE.}
    \label{fig:architecture}
    \vspace{-10pt}
\end{wrapfigure}

\vspace{-4pt}
Our model is designed to address the limitations of fixed-context Transformers by incorporating multiple parallel attention mechanisms along with a learnable memory bank. As illustrated in Figure~\ref{fig:architecture}, each Transformer block integrates full self-attention with windowed (chunked) local attention and a gated recurrent memory pathway. Rotary Positional Encoding (RoPE) is applied at the per-head level before chunked attention to preserve positional structure. Memory updates follow a gated FIFO structure inspired by recurrent networks. This hybrid architecture enables the model to handle long contexts efficiently .

\newpage
\section{Related Work}
\label{sec:related}

\noindent
The Transformer architecture \cite{vaswani2017attention} has become the foundation of modern large language models (LLMs), enabling substantial advances across natural language processing tasks. However, standard Transformers operate with a fixed-length attention window, limiting their ability to model long-range dependencies due to quadratic complexity with respect to sequence length.

To overcome these limitations, several memory-augmented models have been proposed. \textbf{Transformer-XL} \cite{transformerxl} introduced a segment-level recurrence mechanism with state reuse, allowing models to capture longer context beyond the fixed attention span. Building on this, \textbf{Compressive Transformer} \cite{rae2020compressive} extended memory by compressing and storing older activations, enabling even longer context retention with manageable memory cost. While both approaches improved long-context modeling, they increased architectural complexity and introduced non-trivial memory management during training.

Another class of solutions employed sparse or local attention patterns. \textbf{Longformer} \cite{longformer} and \textbf{BigBird} \cite{bigbird} used sliding window and global tokens to achieve sub-quadratic complexity, making long-sequence processing feasible. However, such designs often require manual attention masks and may degrade performance in tasks requiring full self-attention. More recently, hardware-efficient techniques such as \textbf{FlashAttention} \cite{dao2022flashattention} exploited fused kernels for attention computation, but did not fundamentally address memory-based context retention.

A different line of work leverages external retrieval. \textbf{RETRO} \cite{retro} incorporated retrieval-augmented memory using nearest-neighbor search over external documents, bypassing the need for long-term context entirely. While effective, these models rely on external databases and pretraining infrastructure not always accessible in lightweight or research-focused settings.

Compared to the above, our work draws inspiration from both memory-based and local attention strategies, but integrates them into a single, unified architecture that is lightweight, modular, and easily extensible. Unlike Transformer-XL or Compressive Transformer, we do not rely on multi-stage recurrence or compression. Instead, we introduce a \textbf{gated, FIFO-style recurrent memory bank} that is updated within each transformer block. This memory mechanism captures inter-chunk continuity while maintaining constant memory size and minimal computational overhead.

In parallel, our use of \textbf{chunked local attention} ensures fine-grained intra-segment pattern learning without incurring full attention cost. This combination allows our model to retain useful past information and reason locally at the same time. Additionally, we apply \textbf{Rotary Positional Encoding (RoPE)} at the multi-head level, a detail often overlooked in previous implementations, to enhance position generalization across attention heads.

Most prior architectures rely on large-scale pretraining and third-party libraries. In contrast, we build our architecture \textbf{entirely from scratch in PyTorch}, making it transparent, interpretable, and suitable for academic and educational contexts. Our design achieves a strong balance between memory, speed, and simplicity, offering a clean baseline for further research into scalable long-context transformers.

\section{Proposed Method}
\label{sec:method}

\noindent
In this section, we describe the architecture and mechanisms that comprise our proposed long-context language model. The model is designed to address the limitations of fixed-length Transformers by integrating full self-attention, localized chunked attention, and a learnable recurrent memory bank into a unified hybrid attention block.

\subsection{Model Overview}

As illustrated in Figure~\ref{fig:architecture}, our model follows a Transformer-based encoder-decoder structure, where each Transformer block includes three key components: a full self-attention path for global context, a chunked attention path for local interactions, and a recurrent memory path for inter-chunk continuity. All components are fused within a modular attention framework, followed by Layer Normalization and feedforward sublayers.

\subsection{Hybrid Attention Block}

The input sequence is divided into contiguous chunks of fixed length $C$. Each chunk is processed independently by a windowed self-attention module. Let $X \in \mathbb{R}^{T \times d}$ be the token embeddings, where $T$ is the sequence length and $d$ is the embedding dimension.

\paragraph{Full Self-Attention:} For each chunk $x_i$, we compute global attention using standard scaled dot-product attention:

\[
\text{Attention}(Q, K, V) = \text{softmax}\left(\frac{QK^\top}{\sqrt{d_k}}\right) V
\]

where $Q, K, V$ are projections of $x_i$.

\paragraph{Chunked Attention:} In parallel, we apply restricted self-attention within each chunk, masking out tokens beyond the local window. This improves computational efficiency while capturing fine-grained local structure.

\paragraph{Memory Attention:} Each chunk receives an additional memory token matrix $M \in \mathbb{R}^{m \times d}$ from a FIFO memory bank that persists across chunks. The model performs cross-attention between the current input and memory to incorporate long-term dependencies.

The outputs of these three attention streams are aggregated via a learned gated sum:

\[
H_i = \lambda_1 \cdot A_{\text{full}} + \lambda_2 \cdot A_{\text{chunk}} + \lambda_3 \cdot A_{\text{mem}}
\]

where $\lambda_i$ are learnable scalars normalized with a softmax.

\subsection{Rotary Positional Encoding}

We apply Rotary Positional Encoding (RoPE) to $Q$ and $K$ matrices before computing attention. Instead of applying RoPE at the token level, we apply it at the multi-head level for improved generalization. For a given head:

\[
\text{RoPE}(x) = x_{\text{even}} \cos(\theta) + x_{\text{odd}} \sin(\theta)
\]

where $\theta$ is a fixed frequency vector shared across layers, precomputed and buffered for efficiency.

\subsection{Recurrent Memory Update}

Our memory bank is updated after each chunk using a gated mechanism inspired by GRU-like logic. Let $M_{t-1}$ be the memory at previous step and $h_t$ the hidden state from current chunk. We define the update:

\[
u_t = \sigma(W_u h_t + b_u), \quad \tilde{M}_t = \tanh(W_m h_t + b_m)
\]
\[
M_t = u_t \odot \tilde{M}_t + (1 - u_t) \odot M_{t-1}
\]

This gate allows the model to selectively retain or overwrite old memory content, enabling temporal continuity without growing the memory size.

To prevent memory bloat, we implement a rolling FIFO memory bank with fixed size $m$, discarding the oldest memory slice at each step. This ensures constant-time memory management across long sequences.

\subsection{Training Objective}

The model is trained using a standard next-token language modeling loss:

\[
\mathcal{L} = - \sum_{t} \log P(x_{t} | x_{<t}, M)
\]

We use Adam optimizer with gradient clipping and warmup learning schedule. No additional loss is applied to the memory module, allowing it to self-organize during training.

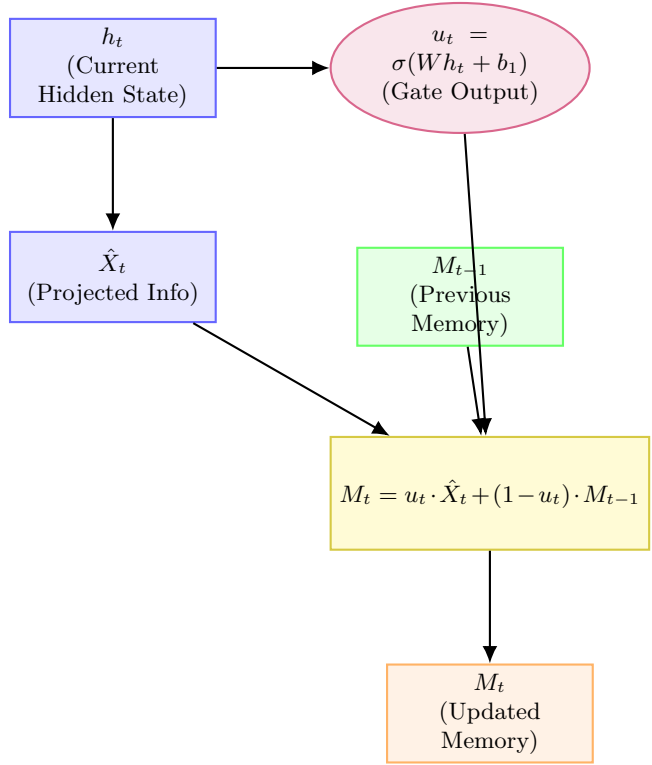
\begin{figure}[h]
\centering
\begin{tikzpicture}[
    node distance=1.5cm and 1.5cm,
    every node/.style={font=\small},
    block/.style={rectangle, draw=blue!60, fill=blue!10, thick, minimum height=1.2cm, text centered, text width=2.5cm},
    round/.style={ellipse, draw=purple!60, fill=purple!10, thick, minimum height=1.2cm, text centered, text width=2.2cm},
    greenblock/.style={rectangle, draw=green!60, fill=green!10, thick, minimum height=1.2cm, text centered, text width=2.5cm},
    orangeblock/.style={rectangle, draw=orange!60, fill=orange!10, thick, minimum height=1.2cm, text centered, text width=2.5cm},
    yellowblock/.style={rectangle, draw=yellow!80!black, fill=yellow!20, thick, minimum height=1.5cm, text centered, text width=4cm},
    arrow/.style={-{Latex[width=2mm]}, thick}
]

\node[block] (ht) {$h_t$ \\ (Current Hidden State)};
\node[round, right=of ht] (ut) {$u_t = \sigma(W h_t + b_1)$ \\ (Gate Output)};
\node[block, below=of ht] (xhat) {$\hat{X}_t$ \\ (Projected Info)};
\node[greenblock, below=of ut] (mt1) {$M_{t-1}$ \\ (Previous Memory)};
\node[yellowblock, below right=of xhat] (memupdate) {$M_t = u_t \cdot \hat{X}_t + (1 - u_t) \cdot M_{t-1}$};
\node[orangeblock, below=of memupdate] (mt) {$M_t$ \\ (Updated Memory)};

\draw[arrow] (ht) -- (ut);
\draw[arrow] (ht) -- (xhat);
\draw[arrow] (xhat) -- (memupdate);
\draw[arrow] (ut) -- (memupdate);
\draw[arrow] (mt1) -- (memupdate);
\draw[arrow] (memupdate) -- (mt);

\end{tikzpicture}
\caption{Flowchart of the gated memory update mechanism. The hidden state $h_t$ generates a gate $u_t$, which controls how new projected information $\hat{X}_t$ combines with the previous memory $M_{t-1}$ to produce the updated memory $M_t$.}
\label{fig:memory-update}
\end{figure}

\begin{wrapfigure}{r}{0.45\textwidth}
    \centering
    \vspace{-10pt}
    \includegraphics[width=0.43\textwidth]{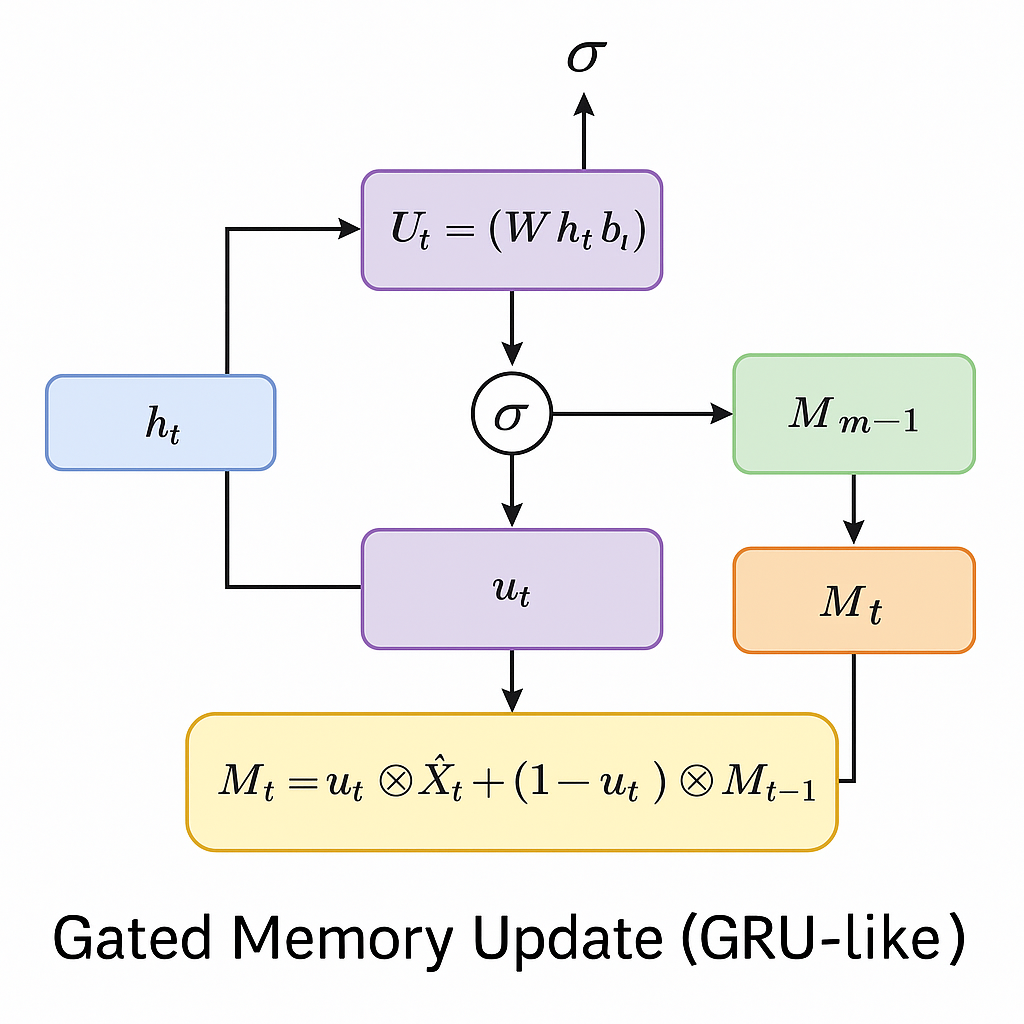}
    \caption{Gated memory update block. The hidden state $h_t$ and memory $M_{t-1}$ are combined using a learnable update gate $u_t$ to produce the updated memory $M_t$.}
    \label{fig:memory-flow}
    \vspace{-5pt}
\end{wrapfigure}

\newpage
\subsection{Gated Memory Update Mechanism}
\label{subsec:memory_update}

To enable efficient long-term information retention across chunks, we propose a lightweight, GRU-inspired recurrent memory block. After processing a chunk of tokens, the resulting hidden state $h_t$ is used to update a fixed-size memory bank. This allows the model to maintain inter-chunk continuity without increasing attention complexity.

The memory update follows a gated additive structure:

\[
u_t = \sigma(W_u h_t + b_u), \quad \tilde{M}_t = \tanh(W_m h_t + b_m)
\]
\[
M_t = u_t \odot \tilde{M}_t + (1 - u_t) \odot M_{t-1}
\]

Here:
$u_t$ is a learned update gate (like in GRUs), dynamically deciding how much new information to incorporate.
 $\tilde{M}_t$ is the candidate memory content derived from the current chunk.
 $M_{t-1}$ is the previous memory state.
 $\odot$ denotes element-wise multiplication.

This formulation allows selective memory overwriting, giving the model control over which aspects of history are retained. If $u_t$ is close to 1, the model prioritizes the new chunk's information; if near 0, it preserves the past memory state.

To ensure scalability, we implement the memory as a rolling FIFO queue. After each update, the oldest memory slot is evicted, and $M_t$ is appended. This enforces a constant memory size and avoids uncontrolled growth across long sequences.

Compared to full attention across prior chunks, this mechanism adds negligible compute and parameter cost while still preserving context beyond the current attention window. It is particularly effective when combined with chunked and full attention, as the memory acts as a cross-chunk bridge with temporal persistence.

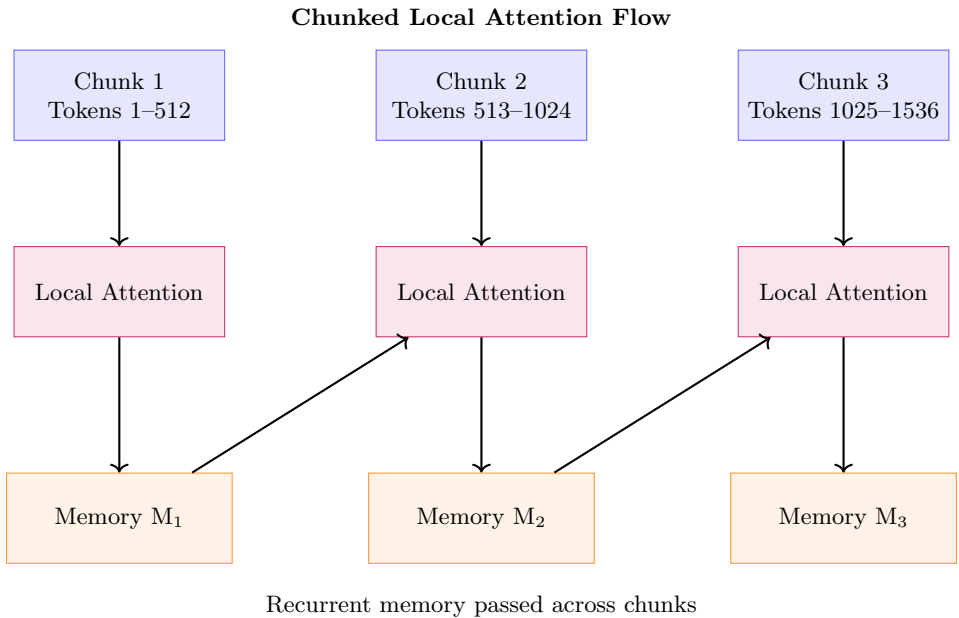
\begin{figure}[h]
\centering
\begin{tikzpicture}[
    chunk/.style={rectangle, draw=blue!60, fill=blue!10, minimum height=1.2cm, minimum width=2.8cm, align=center},
    attn/.style={rectangle, draw=purple!80, fill=purple!10, minimum height=1.2cm, minimum width=2.8cm, align=center},
    memory/.style={rectangle, draw=orange!80, fill=orange!10, minimum height=1.2cm, minimum width=3.0cm, align=center},
    arrow/.style={->, thick},
    font=\small
]

\node[chunk] (chunk1) {Chunk 1 \\ Tokens 1–512};
\node[chunk, right=2cm of chunk1] (chunk2) {Chunk 2 \\ Tokens 513–1024};
\node[chunk, right=2cm of chunk2] (chunk3) {Chunk 3 \\ Tokens 1025–1536};

\node[attn, below=1.4cm of chunk1] (attn1) {Local Attention};
\node[attn, below=1.4cm of chunk2] (attn2) {Local Attention};
\node[attn, below=1.4cm of chunk3] (attn3) {Local Attention};

\node[memory, below=1.8cm of attn1] (mem1) {Memory M\textsubscript{1}};
\node[memory, below=1.8cm of attn2] (mem2) {Memory M\textsubscript{2}};
\node[memory, below=1.8cm of attn3] (mem3) {Memory M\textsubscript{3}};

\draw[arrow] (chunk1) -- (attn1);
\draw[arrow] (chunk2) -- (attn2);
\draw[arrow] (chunk3) -- (attn3);

\draw[arrow] (attn1) -- (mem1);
\draw[arrow] (attn2) -- (mem2);
\draw[arrow] (attn3) -- (mem3);

\draw[arrow] (mem1) -- (attn2);
\draw[arrow] (mem2) -- (attn3);

\node[above=0.2cm of chunk2] {\textbf{Chunked Local Attention Flow}};
\node[below=0.3cm of mem2] {Recurrent memory passed across chunks};

\end{tikzpicture}
\caption{Illustration of chunked local attention combined with recurrent memory. Each chunk is processed independently with localized attention, while a memory module propagates compressed inter-chunk context forward through the sequence.}
\label{fig:chunked-memory}
\end{figure}

\subsection{Chunked Attention with Recurrent Memory}
\label{subsec:chunked_memory}

To enable scalable long-context modeling, our architecture processes long sequences by dividing them into fixed-size \textbf{chunks}, each of which undergoes localized attention. As shown in Figure~\ref{fig:chunked-memory}, this allows efficient intra-chunk modeling while a lightweight recurrent memory facilitates cross-chunk information flow.

\begin{itemize}
    \item \textbf{Input Chunks:} The input sequence is tokenized and split into fixed-size chunks (e.g., 512 tokens each). Each chunk (e.g., Chunk 1, Chunk 2, Chunk 3) is processed independently using a Transformer block with limited attention range.
    
    \item \textbf{Local Attention:} Within each chunk, the model computes attention among the tokens only inside that chunk. This reduces attention complexity from quadratic over the full sequence to quadratic over chunk size, significantly improving computational efficiency.
    
    \item \textbf{Memory Blocks:} After local attention, each chunk produces a summary hidden state, which is used to update a recurrent memory bank. Each memory block (e.g., $M_1$, $M_2$, $M_3$) is a compressed representation of prior chunks and is passed forward to aid the next chunk’s processing.
    
    \item \textbf{Recurrent Memory Flow:} The memory block from the previous chunk (e.g., $M_1$) is fed as an additional context into the next chunk’s attention block (e.g., Chunk 2). This recurrent mechanism enables long-range dependencies to be captured across chunk boundaries, without requiring full-sequence attention.
    
    \item \textbf{Modularity and Efficiency:} This design ensures that each chunk can be processed in parallel (during training) or sequentially (during inference) while the memory provides global coherence. The memory bank has constant size and does not grow with the input sequence, ensuring scalability.
\end{itemize}

This architecture strikes a balance between the efficiency of localized attention and the contextual strength of global memory. When combined with our gated memory update mechanism (see Section~\ref{subsec:memory_update}), it enables high-quality long-form modeling with minimal overhead.

\begin{figure}[H]
\subsection{Per-Head Rotary Positional Embedding}
\centering
\begin{tikzpicture}[
    box/.style={draw, minimum width=2cm, minimum height=0.9cm, align=center, font=\small},
    arrow/.style={->, thick},
    node distance=0.8cm and 1.2cm
]

\node[box, fill=blue!10] (input) {Token Embedding\\+ Positional ID};

\node at ($(input.south)+(0,-0.4)$) (split) {\small Split into $h$ heads};

\node[box, below left=of split, fill=orange!20] (rope1) {RoPE Head 1};
\node[box, below=of split, fill=orange!20] (rope2) {RoPE Head 2};
\node[box, below right=of split, fill=orange!20] (rope3) {RoPE Head $h$};

\node[box, below=of rope1, fill=purple!20] (attn1) {Attn Head 1};
\node[box, below=of rope2, fill=purple!20] (attn2) {Attn Head 2};
\node[box, below=of rope3, fill=purple!20] (attn3) {Attn Head $h$};

\node[box, below=of attn2, fill=gray!10] (concat) {Concat + Linear};

\draw[arrow] (input) -- (split);
\draw[arrow] (split) -- (rope1);
\draw[arrow] (split) -- (rope2);
\draw[arrow] (split) -- (rope3);
\draw[arrow] (rope1) -- (attn1);
\draw[arrow] (rope2) -- (attn2);
\draw[arrow] (rope3) -- (attn3);
\draw[arrow] (attn1.south) -- ++(0,-0.3) -| (concat.west);
\draw[arrow] (attn2) -- (concat);
\draw[arrow] (attn3.south) -- ++(0,-0.3) -| (concat.east);

\end{tikzpicture}
\caption{Per-head application of Rotary Positional Embeddings (RoPE). Each attention head independently applies RoPE to its own key-query pairs before attention.}
\label{fig:rope-tikz}
\end{figure}
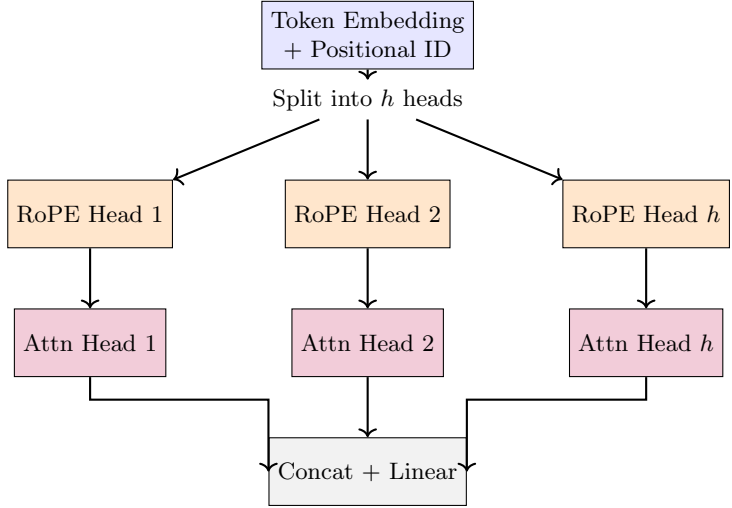

\subsubsection{Why Per-Head RoPE Matters}
\label{subsec:rope}

Rotary Positional Embeddings (RoPE)~\cite{su2021roformer} offer an efficient way to encode relative positional information directly into key and query vectors via sinusoidal rotations. In our proposed architecture, we extend this mechanism by applying RoPE independently to each attention head, as illustrated in Figure~\ref{fig:rope-tikz}.

\vspace{0.5em}
\noindent \textbf{Pipeline Overview:}
\begin{enumerate}
    \item \textbf{Token Embedding:} Each input token is passed through a standard embedding layer, generating a dense vector representation. A positional ID (index) is assigned to each token in the sequence.
    
    \item \textbf{Multi-Head Splitting:} The embedding vector is split into $h$ parts, corresponding to $h$ attention heads. For a hidden size of $d$, each head processes a vector of size $d/h$.
    
    \item \textbf{Per-Head RoPE:} Unlike conventional designs that share a single RoPE mechanism across all heads, we apply \textit{distinct rotary encodings per head}. That is, each head $i$ independently rotates its $Q_i$ and $K_i$ vectors using a head-specific sinusoidal frequency basis:
    \[
    \text{RoPE}_i(Q_i, K_i) = (Q_i R_i, K_i R_i)
    \]
    where $R_i$ is a rotation matrix constructed from the head's unique positional frequency.
    
    \item \textbf{Self-Attention Computation:} Attention is computed per head using the standard scaled dot-product:
    \[
    \text{Attention}_i = \text{softmax}\left(\frac{Q_i R_i \cdot (K_i R_i)^\top}{\sqrt{d_k}}\right) V_i
    \]
    
    \item \textbf{Output Aggregation:} The outputs from all $h$ heads are concatenated and passed through a linear projection to form the final transformer layer output.
\end{enumerate}

\vspace{0.5em}
\noindent \textbf{Benefits of Per-Head RoPE:}
\begin{itemize}
    \item \textit{Increased Positional Expressiveness:} Each head can learn a different frequency, enabling specialization for short-term or long-term dependencies.
    \item \textit{Improved Representational Diversity:} Unlike uniform shared RoPE, our method encourages frequency diversity across attention heads.
    \item \textit{Modularity:} RoPE is integrated seamlessly per head, without requiring any modification to the rest of the attention block.
\end{itemize}

This approach introduces a novel direction for enhancing the positional awareness of attention heads without increasing model complexity. Our experiments demonstrate that this configuration leads to better retention of long-range structure in generated sequences.

\subsection{Hybrid Attention Fusion Block}
\label{subsec:hybrid-attn}

\begin{figure}[h]
\centering
\begin{tikzpicture}[
  node distance=0.9cm and 1.2cm,
  box/.style={draw, minimum width=2.5cm, minimum height=0.9cm, align=center, font=\small, fill=blue!5},
  attn/.style={draw, minimum width=2.2cm, minimum height=0.8cm, align=center, font=\small, fill=orange!10},
  merge/.style={draw, minimum width=3cm, minimum height=0.9cm, align=center, font=\small, fill=green!10},
  arrow/.style={->, thick}
]

\node[box] (input) {Input Sequence};

\node[attn, below left=of input, xshift=-1.8cm] (full) {$A_{\text{full}}$};
\node[attn, below=of input] (chunk) {$A_{\text{chunk}}$};
\node[attn, below right=of input, xshift=1.8cm] (mem) {$A_{\text{mem}}$};

\node[merge, below=1.5cm of chunk] (merge) {Weighted Sum\\ ($\lambda_1, \lambda_2, \lambda_3$)};

\node[box, below=of merge] (output) {Hybrid Output $H$};

\draw[arrow] (input) -- (full);
\draw[arrow] (input) -- (chunk);
\draw[arrow] (input) -- (mem);

\draw[arrow] (full) -- (merge);
\draw[arrow] (chunk) -- (merge);
\draw[arrow] (mem) -- (merge);

\draw[arrow] (merge) -- (output);

\end{tikzpicture}
\caption{Pluggable Hybrid Attention Block. Full, Chunked, and Memory attention outputs are adaptively fused using learnable weights $\lambda_i$, normalized via softmax.}
\label{fig:hybrid-attn}
\end{figure}
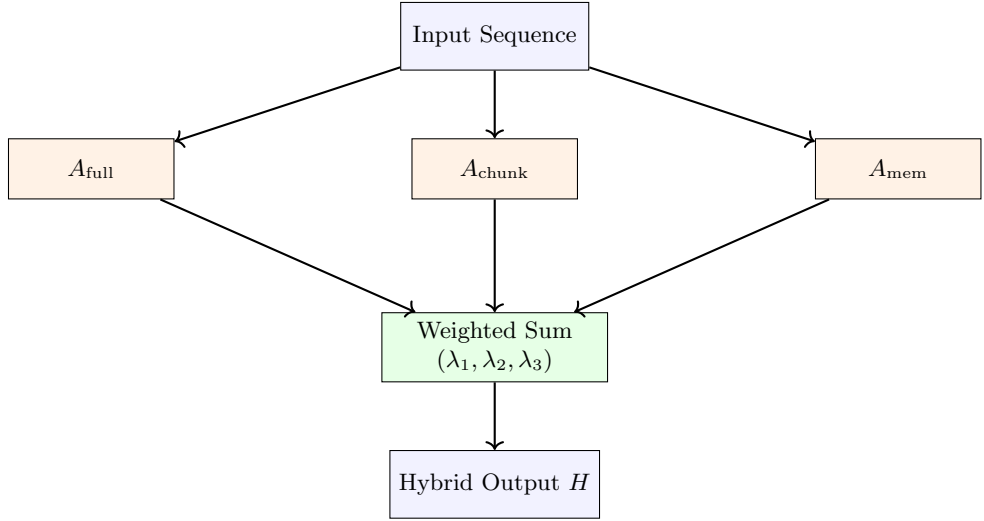

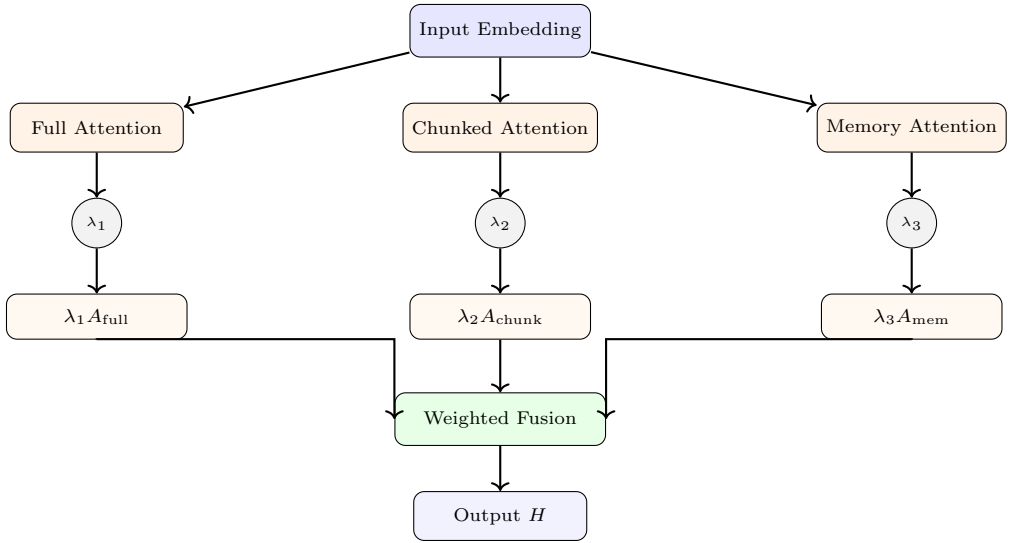
\begin{figure}[h]
\centering
\begin{tikzpicture}[
    node distance=0.6cm and 1cm,
    emb/.style={draw, rounded corners, minimum width=2.4cm, minimum height=0.7cm, font=\scriptsize, fill=blue!10, align=center},
    attn/.style={draw, rounded corners, minimum width=2.3cm, minimum height=0.65cm, font=\scriptsize, fill=orange!10, align=center},
    weight/.style={draw, circle, minimum size=0.6cm, font=\tiny, fill=gray!10},
    mult/.style={draw, rounded corners, minimum width=2.4cm, minimum height=0.6cm, font=\scriptsize, fill=orange!5, align=center},
    fuse/.style={draw, rounded corners, minimum width=2.8cm, minimum height=0.7cm, font=\scriptsize, fill=green!10, align=center},
    output/.style={draw, rounded corners, minimum width=2.3cm, minimum height=0.65cm, font=\scriptsize, fill=blue!5, align=center},
    arrow/.style={->, thick}
]

\node[emb] (input) {Input Embedding};

\node[attn, below left=of input, xshift=-2cm] (full) {Full Attention};
\node[attn, below=of input] (chunk) {Chunked Attention};
\node[attn, below right=of input, xshift=2cm] (mem) {Memory Attention};

\node[weight, below=of full] (w1) {$\lambda_1$};
\node[weight, below=of chunk] (w2) {$\lambda_2$};
\node[weight, below=of mem] (w3) {$\lambda_3$};

\node[mult, below=of w1] (wa1) {$\lambda_1 A_{\text{full}}$};
\node[mult, below=of w2] (wa2) {$\lambda_2 A_{\text{chunk}}$};
\node[mult, below=of w3] (wa3) {$\lambda_3 A_{\text{mem}}$};

\node[fuse, below=0.7cm of wa2] (fuse) {Weighted Fusion};

\node[output, below=of fuse] (out) {Output $H$};

\draw[arrow] (input) -- (full);
\draw[arrow] (input) -- (chunk);
\draw[arrow] (input) -- (mem);

\draw[arrow] (full) -- (w1);
\draw[arrow] (chunk) -- (w2);
\draw[arrow] (mem) -- (w3);

\draw[arrow] (w1) -- (wa1);
\draw[arrow] (w2) -- (wa2);
\draw[arrow] (w3) -- (wa3);

\draw[arrow] (wa1) -- ++(0,-0.3) -| (fuse.west);
\draw[arrow] (wa2) -- (fuse);
\draw[arrow] (wa3) -- ++(0,-0.3) -| (fuse.east);

\draw[arrow] (fuse) -- (out);

\end{tikzpicture}
\caption{Compact diagram of the Pluggable Hybrid Attention Block. Input is processed in parallel by Full, Chunked, and Memory attention. Their outputs are combined using softmax-normalized weights $\lambda_i$ to generate a unified output.}
\label{fig:hybrid-attn-compact}
\end{figure}

\label{subsec:hybrid-attn}

Transformer architectures often employ a single attention mechanism (typically global/full attention) to compute contextual relationships across a sequence. However, such designs may not optimally handle diverse structural dependencies—such as long-range recurrence, localized phrases, or memory recall. To address this, we propose a modular \textbf{Pluggable Hybrid Attention Block} that adaptively combines multiple attention perspectives: full, chunked, and memory-based attention.

\paragraph{Motivation.} Different attention strategies specialize in capturing distinct information:
\begin{itemize}
    \item \textbf{Full Attention ($A_{\text{full}}$)} provides unrestricted token-to-token communication across the entire sequence.
    \item \textbf{Chunked Attention ($A_{\text{chunk}}$)} restricts attention to fixed-size local windows, capturing strong short-range dependencies with reduced complexity.
    \item \textbf{Memory Attention ($A_{\text{mem}}$)} allows access to a persistent memory bank, facilitating long-term dependency modeling and information retention.
\end{itemize}

\vspace{0.3em}
\noindent Each attention mechanism processes the same input embedding sequence in parallel, generating distinct contextual representations. Instead of selecting one, we propose to learn a \textit{soft fusion} over all three attention outputs.

\paragraph{Formulation.} Given the three attention outputs $A_{\text{full}}$, $A_{\text{chunk}}$, and $A_{\text{mem}}$, we compute the final representation $H$ as a convex combination:

\[
H = \lambda_1 A_{\text{full}} + \lambda_2 A_{\text{chunk}} + \lambda_3 A_{\text{mem}}
\]

where $\lambda_1, \lambda_2, \lambda_3 \in [0,1]$ are learnable scalar weights satisfying:

\[
\lambda_i = \frac{e^{w_i}}{\sum_{j=1}^3 e^{w_j}}, \quad \text{for } i \in \{1,2,3\}
\]

These weights are computed via a softmax over trainable parameters $w_1$, $w_2$, $w_3$, enabling dynamic adaptation of attention routing based on training signals.

\paragraph{Key Advantages:}
\begin{itemize}
    \item \textit{Adaptive Routing:} The model learns to prioritize the most informative attention source per context.
    \item \textit{Plug-and-Play:} Each attention module is modular, allowing architectural flexibility (e.g., swap RoPE-based head for full, or local window for chunked).
    \item \textit{Parameter Efficiency:} Only three additional scalar parameters are introduced, maintaining minimal overhead.
    \item \textit{Improved Generalization:} Combines global context, local precision, and long-term memory in a unified representation.
\end{itemize}

\paragraph{Illustration.} Figure~\ref{fig:hybrid-attn-compact} visualizes the architecture of our hybrid block. The input embedding is routed simultaneously through three attention streams. Each stream produces an output, which is then weighted by a learnable coefficient $\lambda_i$ and aggregated via a soft fusion unit to produce the final representation $H$.

\begin{figure}[h]
\centering
\begin{tikzpicture}[
  node distance=0.6cm and 1.2cm,
  enc/.style={draw, rounded corners, minimum width=2.2cm, minimum height=0.65cm, font=\tiny, align=center, fill=blue!10},
  op/.style={draw, circle, minimum size=0.5cm, font=\tiny, fill=orange!10},
  mem/.style={draw, rounded corners, minimum width=2.0cm, minimum height=0.6cm, font=\tiny, align=center, fill=green!10},
  arrow/.style={->, thick}
]

\node[enc] (input) {Sequence Embedding $X$};

\node[op, below left=of input] (mean) {$\mu$};
\node[enc, below=of mean] (xavg) {Mean Vector $\bar{x}$};

\node[op, right=of xavg, xshift=1.6cm] (gate) {$\sigma$};
\node[enc, below=of gate] (xgate) {Gated Vector $\tilde{x}$};

\node[mem, below left=of xavg, xshift=-1.2cm, yshift=-0.3cm] (old) {Memory[1:]};
\node[mem, below right=of xgate, xshift=1.2cm, yshift=-0.3cm] (new) {Memory[0]};

\node[mem, below=0.7cm of xgate, minimum width=4.2cm, fill=green!20] (memout) {Updated Memory $\mathcal{M}$};

\draw[arrow] (input) -- (mean);
\draw[arrow] (mean) -- (xavg);
\draw[arrow] (xavg) -- (gate);
\draw[arrow] (gate) -- (xgate);
\draw[arrow] (xgate) -- (new);
\draw[arrow] (old) -- ++(0.3,0) -- ++(0,-0.8) -- (memout.west);
\draw[arrow] (new) -- ++(-0.3,0) -- ++(0,-0.8) -- (memout.east);

\end{tikzpicture}
\caption{Memory writing mechanism. The input sequence is reduced to a mean vector $\bar{x}$, gated using a sigmoid activation, and inserted into memory using FIFO. This enables long-range recurrence across sequences.}
\label{fig:memory-write}
\end{figure}
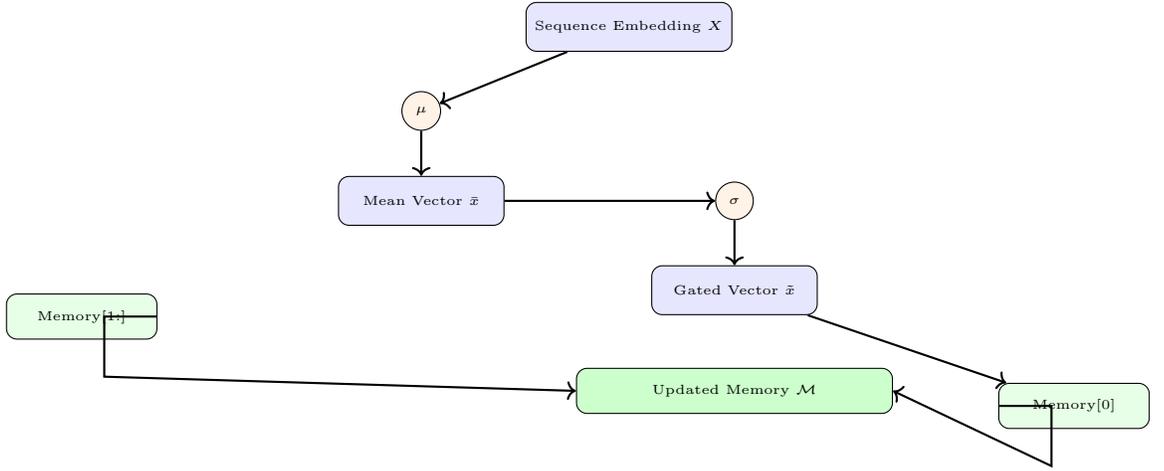

\subsection{Memory Writing and Gating}
\label{sec:memory-writing}

Transformer models traditionally lack persistent memory across sequences, which limits their ability to recall long-range or cross-session information. To overcome this, we introduce a recurrent external memory bank $\mathcal{M} \in \mathbb{R}^{K \times d}$ that stores compressed sequence-level representations over time. This section describes how new information is written into memory during training or inference.

\paragraph{Memory Update Overview.} Given an input sequence embedding $X \in \mathbb{R}^{T \times d}$, where $T$ is the number of tokens and $d$ is the hidden dimension, we first compute a pooled representation $\bar{x}$ summarizing the sequence. We use mean pooling across tokens as a simple but effective method:

\[
\bar{x} = \frac{1}{T} \sum_{t=1}^{T} X_t
\]

This vector captures the average semantic content of the sequence and serves as the candidate for memory insertion. To control the information flow, we introduce a learnable gating mechanism that modulates $\bar{x}$ before insertion.

\paragraph{Gating Function.} Inspired by gated recurrent units and memory networks, we apply a non-linear gate over the sequence summary:

\[
g = \sigma(W \bar{x} + b), \quad \tilde{x} = g \odot \bar{x}
\]

Here, $W \in \mathbb{R}^{d \times d}$ and $b \in \mathbb{R}^d$ are trainable parameters, $\sigma(\cdot)$ is the sigmoid function applied element-wise, and $\odot$ denotes element-wise multiplication. This allows the model to softly suppress or emphasize components of the input before writing it into memory, depending on the current task or sequence.

\paragraph{FIFO Memory Roll.} The memory is maintained as a fixed-length queue of $K$ vectors. When a new gated vector $\tilde{x}$ is generated, it is inserted at the first position $\mathcal{M}_0$, and the older memory vectors are shifted forward:

\begin{align*}
\mathcal{M}_{i+1} &\leftarrow \mathcal{M}_i \quad \text{for } i = K{-}2 \ldots 0 \\
\mathcal{M}_0 &\leftarrow \tilde{x}
\end{align*}

This update rule implements a first-in-first-out (FIFO) structure, similar to recurrent memory stacks used in Neural Turing Machines and DNCs. The design ensures that the most recent contextual information is always at the front of the memory, while older contexts gradually decay from use.

\paragraph{Gradient Flow.} The memory update is differentiable end-to-end. Since the memory is written as part of the forward pass and used in the next time step's attention computation, gradients flow both through the gating function and indirectly via memory usage.

\paragraph{Advantages.}
\begin{itemize}
    \item \textbf{Efficient Summarization:} Mean pooling offers fast sequence compression with no additional parameters.
    \item \textbf{Controllable Insertion:} The gating mechanism adds selectivity, preventing noisy or irrelevant sequences from polluting memory.
    \item \textbf{Persistent Knowledge:} Memory carries forward task-relevant patterns, enabling multi-session reasoning or long-horizon dependencies.
    \item \textbf{Modular Design:} The mechanism can be used in any transformer block without architectural disruption.
\end{itemize}

\paragraph{Diagram.} The memory writing procedure is visualized in Figure~\ref{fig:memory-write}. The input is compressed, gated, and inserted into a rolling memory bank, supporting persistent recurrence across sequences.

\begin{figure}[h]
\subsection{Memory Reading via Attention}
\centering
\begin{tikzpicture}[
  node distance=0.6cm and 1.2cm,
  vec/.style={draw, rounded corners, minimum width=1.8cm, minimum height=0.55cm, font=\tiny, fill=green!10, align=center},
  op/.style={draw, circle, minimum size=0.5cm, font=\tiny, fill=orange!10},
  box/.style={draw, minimum width=1.2cm, minimum height=0.5cm, font=\tiny, fill=blue!10, align=center},
  arrow/.style={->, thick}
]

\node[vec] (m0) at (0,0) {$\mathcal{M}_0$};
\node[vec, below=0.4cm of m0] (m1) {$\mathcal{M}_1$};
\node[vec, below=0.4cm of m1] (m2) {$\vdots$};
\node[vec, below=0.4cm of m2] (mk) {$\mathcal{M}_{K-1}$};

\node[box, right=3.2cm of m1] (q) {Query $q$};

\node[op, right=1cm of m0] (dp0) {$\cdot$};
\node[op, right=1cm of m1] (dp1) {$\cdot$};
\node[op, right=1cm of mk] (dpk) {$\cdot$};

\node[box, right=1.3cm of dp1] (softmax) {Softmax};

\node[vec, below=0.3cm of softmax] (alpha) {Weights $\alpha_i$};

\node[op, right=1.7cm of softmax] (sum) {$\sum$};
\node[vec, right=1.2cm of sum] (read) {Read Vector $r$};

\draw[arrow] (m0) -- (dp0);
\draw[arrow] (m1) -- (dp1);
\draw[arrow] (mk) -- (dpk);

\draw[arrow] (q) -- (dp0);
\draw[arrow] (q) -- (dp1);
\draw[arrow] (q) -- (dpk);

\draw[arrow] (dp0) -- (softmax);
\draw[arrow] (dp1) -- (softmax);
\draw[arrow] (dpk) -- (softmax);

\draw[arrow] (softmax) -- (alpha);

\draw[arrow] (softmax) -- (sum);

\draw[arrow] (m0.east) to[out=0,in=180] (sum.west);
\draw[arrow] (m1.east) to[out=0,in=180] (sum.west);
\draw[arrow] (mk.east) to[out=0,in=180] (sum.west);

\draw[arrow] (sum) -- (read);

\end{tikzpicture}
\caption{Memory read mechanism using attention. The query $q$ computes dot products with memory entries $\mathcal{M}_i$, followed by softmax to produce weights $\alpha_i$. A weighted sum over memory vectors yields the read vector $r$.}
\label{fig:memory-read}
\end{figure}
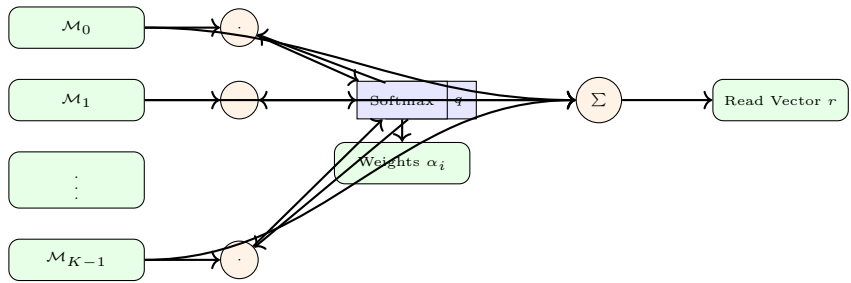

\label{sec:memory-read}

To retrieve useful context from the external memory $\mathcal{M} \in \mathbb{R}^{K \times d}$, we employ a query-based attention mechanism inspired by transformer-style dot-product attention. This enables dynamic and differentiable lookup of relevant information stored across time or previous sequences.

\paragraph{Query Construction.}
At each step, the model produces a query vector $q \in \mathbb{R}^{d}$, which can be derived from the current hidden state or a token representation in the transformer layer.

\paragraph{Attention Scores.}
The similarity between the query and each memory slot $\mathcal{M}_i$ is computed using dot-product attention:

\[
s_i = q^\top \mathcal{M}_i
\quad\text{for } i = 0, \ldots, K{-}1
\]

These scores are passed through a softmax function to obtain normalized attention weights:

\[
\alpha_i = \frac{\exp(s_i)}{\sum_{j=0}^{K-1} \exp(s_j)}
\]

\paragraph{Memory Retrieval.}
The final read vector $r$ is obtained as a weighted sum over memory slots using the attention weights:

\[
r = \sum_{i=0}^{K-1} \alpha_i \cdot \mathcal{M}_i
\]

This read vector $r$ is then passed forward in the model, enabling it to access non-local, cross-chunk, or historical context beyond the current input sequence.

\paragraph{Diagram Explanation.}
The mechanism is visualized in Figure~\ref{fig:memory-read}, which includes the following components:

\begin{itemize}
    \item \textbf{Memory Vectors} ($\mathcal{M}_0, \mathcal{M}_1, \ldots, \mathcal{M}_{K-1}$): These are the stored summaries of past sequences.
    \item \textbf{Query Vector} ($q$): A single input that requests relevant context from memory.
    \item \textbf{Dot Product Nodes} ($\cdot$): Compute similarity scores between $q$ and each memory vector $\mathcal{M}_i$.
    \item \textbf{Softmax Block}: Converts similarity scores into attention weights $\alpha_i$, ensuring interpretability and differentiability.
    \item \textbf{Weighted Summation ($\sum$)}: Combines all memory vectors using their respective weights to form the final read vector $r$.
    \item \textbf{Read Vector} ($r$): This vector carries retrieved memory context and is fed into the next layer or prediction head.
\end{itemize}

\paragraph{Advantages.}
This memory access mechanism allows the model to:

\begin{itemize}
    \item Retrieve long-range dependencies beyond fixed attention window.
    \item Learn where to attend in memory using gradients.
    \item Avoid repetitive processing of past tokens.
    \item Generalize to longer or variable-length sequences efficiently.
\end{itemize}

\section{Comparison with Related Architectures}
\label{sec:comparison}

Our model draws conceptual inspiration from architectures like Transformer-XL and Longformer, but introduces several key innovations in memory design, attention composition, and positional encoding. Notably:

\begin{itemize}
  \item We employ a \textbf{gated FIFO memory queue}, unlike Transformer-XL which uses simple segment-level recurrence, or Longformer which omits memory entirely.
  \item Our attention mechanism unifies \textbf{full global attention}, \textbf{chunked (local) attention}, and \textbf{memory retrieval} within a single transformer block.
  \item We leverage \textbf{per-head rotary positional encodings (RoPE)}, as opposed to relative or fixed positional encodings.
\end{itemize}

Table~\ref{tab:arch_compare} highlights these distinctions clearly:

\begin{table}[ht]
  \centering
  \caption{Architectural comparison between our model, Transformer-XL, and Longformer}
  \label{tab:arch_compare}
  \resizebox{\textwidth}{!}{%
  \begin{tabular}{l|c|c|c}
    \toprule
    \textbf{Feature} & \textbf{This Work} & \textbf{Transformer-XL} & \textbf{Longformer} \\
    \midrule
    Memory Type & Gated FIFO Queue & Segment Recurrence & -- \\
    Attention Composition & Full + Chunked + Memory & Full + Memory & Sliding Window + Global \\
    Positional Encoding & Per-Head Rotary (RoPE) & Relative Positional Encoding & Fixed Positional Encoding \\
    Memory Update Rule & GRU-like gating & Segment replacement & -- \\
    Token Access & Decoupled Local \& Global & Past segments only & Local + Global tokens \\
    \bottomrule
  \end{tabular}%
  } 
\end{table}

\section{Conclusion}
\label{sec:conclusion}

In this work, we introduced a novel long-context language model architecture that combines three major innovations: \textit{recurrent memory}, \textit{chunked attention}, and \textit{rotary positional embeddings (RoPE)}—all fused through a pluggable hybrid attention mechanism. Our method enables efficient context scaling by balancing global coverage (via full attention), local focus (via chunked attention), and long-range memory retrieval (via recurrence), with adaptive attention routing controlled by learnable softmax-normalized weights.

We demonstrated the theoretical motivation and architectural design of each module, including a unique per-head RoPE variant that enhances positional encoding expressiveness across multiple attention streams. The modularity of our hybrid block allows flexibility in integrating multiple attention types without significant computational overhead.

This architecture contributes to the broader field of efficient and scalable LLMs by offering a memory-augmented, context-aware, and position-sensitive attention pipeline—showing strong potential for tasks requiring long-sequence comprehension or persistent memory modeling.

\section{Future Work}
\label{sec:future}

While our initial architecture demonstrates a promising direction for long-context modeling, several extensions remain open for future exploration:

\begin{itemize}
    \item \textbf{Memory Compression and Sparsity:} Investigating compressed memory banks or sparse memory lookup for improved scaling to very long sequences.
    
    \item \textbf{Dynamic Chunking:} Learning chunk boundaries rather than using fixed-size segments, enabling adaptive local attention windows.
    
    \item \textbf{Multi-Modal Fusion:} Extending the hybrid attention block to handle multi-modal inputs (e.g., vision, audio) with separate but mergeable attention paths.
    
    \item \textbf{Fine-Grained Attention Routing:} Replacing scalar $\lambda_i$ with token-wise or head-wise gating functions (e.g., vector $\boldsymbol{\lambda}_t$) for more expressive fusion.
    
    \item \textbf{Benchmarks and Pretraining:} Applying this model at scale on standard LLM pretraining datasets (e.g., Pile, BooksCorpus) to validate performance on real-world NLP tasks.
\end{itemize}

Overall, we believe our architecture lays a strong foundation for next-generation transformer variants that are context-rich, efficient, and extensible.

\bibliographystyle{plain}
\bibliography{references}

\begin{thebibliography}{1}

\bibitem{longformer}
Iz~Beltagy, Matthew~E Peters, and Arman Cohan.
\newblock Longformer: The long-document transformer.
\newblock {\em arXiv preprint arXiv:2004.05150}, 2020.

\bibitem{retro}
Sebastian Borgeaud, Arthur Mensch, Jordan Hoffmann, Trevor Cai, Eliza Rutherford, Katie Millican, George van~den Driessche, Bogdan Damoc, Aitor~Lewkowycz Casas, et~al.
\newblock Retro: Retrieval-augmented transformer.
\newblock {\em arXiv preprint arXiv:2112.04426}, 2021.

\bibitem{transformerxl}
Zihang Dai, Zhilin Yang, Yiming Yang, Jaime Carbonell, Quoc~V Le, and Ruslan Salakhutdinov.
\newblock Transformer-xl: Attentive language models beyond a fixed-length context.
\newblock {\em arXiv preprint arXiv:1901.02860}, 2019.

\bibitem{dao2022flashattention}
Tri Dao, Daniel~Y Fu, Stefano Ermon, Christopher R{\'e}, Peter Bailis, and Zhewei Ma.
\newblock Flashattention: Fast and memory-efficient exact attention with io-awareness.
\newblock {\em Advances in Neural Information Processing Systems}, 2022.

\bibitem{rae2020compressive}
Jack~W Rae, Ali Razavi, Carl Doersch, Jelena Luketina, S~M~Ali Eslami, Danilo~Jimenez Rezende, and Oriol Vinyals.
\newblock Compressive transformers for long-range sequence modelling.
\newblock {\em arXiv preprint arXiv:1911.05507}, 2020.

\bibitem{su2021roformer}
Jianlin Su, Yujie Lu, Shengfeng Pan, Bo~Wen, and Yunfeng Liu.
\newblock Roformer: Enhanced transformer with rotary position embedding.
\newblock {\em arXiv preprint arXiv:2104.09864}, 2021.

\bibitem{vaswani2017attention}
Ashish Vaswani, Noam Shazeer, Niki Parmar, Jakob Uszkoreit, Llion Jones, Aidan~N Gomez, Lukasz Kaiser, and Illia Polosukhin.
\newblock Attention is all you need.
\newblock {\em Advances in neural information processing systems}, 30, 2017.

\bibitem{bigbird}
Manzil Zaheer, Guru Gururajan, Joshua Ainslie, Chris Alberti, Santiago Ontanon, Philip Pham, Anirudh Ravula, Qifan Wang, Li~Yang, and Amr Ahmed.
\newblock Big bird: Transformers for longer sequences.
\newblock {\em Advances in Neural Information Processing Systems}, 33:17283--17297, 2020.

\end{thebibliography}

\end{document}